%% file: 0-Main.tex
\def\year{2021}\relax
\documentclass[letterpaper]{article} 
\usepackage{aaai21}  
\usepackage{times}  
\usepackage{helvet} 
\usepackage{courier}  
\usepackage[hyphens]{url}  
\usepackage{graphicx} 
\usepackage{natbib}  
\usepackage{caption} 
\usepackage{algorithm}
\usepackage{algorithmic}
\usepackage{multirow}
\usepackage{booktabs}
\usepackage{diagbox}
\usepackage{subfigure}
\usepackage{amsmath,amsfonts,amssymb}
\usepackage{bm}

\newtheorem{definition}{\bf Definition}
\title{Overcoming Catastrophic Forgetting in Graph Neural Networks\\ with Experience Replay}
\author{

    Fan Zhou, 
    Chengtai Cao\textsuperscript{\rm *}
}
\affiliations{

    University of Electronic Science and Technology of China\\

    fan.zhou@uestc.edu.cn, cct447131988@gmail.com\\
    \textsuperscript{\rm *} Corresponding author.

}

\begin{document}

\maketitle

\begin{abstract}
Graph Neural Networks (GNNs) have recently received significant research attention due to their superior performance on a variety of graph-related learning tasks. Most of the current works focus on either static or dynamic graph settings, addressing a single particular task, e.g., node/graph classification, link prediction. In this work, we investigate the question: can GNNs be applied to \textit{continuously} learning a sequence of tasks? Towards that, we explore the Continual Graph Learning (CGL) paradigm and present the \textbf{\underline{E}}xperience \textbf{\underline{R}}eplay based framework \textbf{ER-GNN} for CGL to alleviate the catastrophic forgetting problem in existing GNNs. ER-GNN stores knowledge from previous tasks as experiences and replays them when learning new tasks to mitigate the catastrophic forgetting issue. We propose three experience node selection strategies: \textit{mean of feature}, \textit{coverage maximization}, and \textit{influence maximization}, to guide the process of selecting experience nodes. Extensive experiments on three benchmark datasets demonstrate the effectiveness of our ER-GNN and shed light on the incremental graph (non-Euclidean) structure learning.
\end{abstract}

\section{Introduction}
\label{introduction}
\input{1_Introduction.tex}

\section{Related Work}
\label{RelatedWork}
\input{2_RelatedWork.tex}

\section{Methodology}
\label{Methodology}
\input{3_Methodology.tex}

\section{Experiments}
\label{experiments}
\input{4_Experiments.tex}

\section{Conclusion}
\label{Conclusion}
In this work, we formulated a novel practical graph-based continual learning problem, where the GNN model is expected to learn a sequence of node classification tasks without catastrophic forgetting. We presented a general framework called ER-GNN that exploits experience replay based methods to mitigate the impacts of forgetting. We also discussed three experience selection schemes, including a novel one -- IM (Influence Maximization), which utilizes influence function to select experience nodes. The extensive experiments demonstrated the effectiveness and applicability of our ER-GNN. As part of our future, we plan to extend the continual learning to different graph-related tasks, such as graph alignment and cascade/diffusion prediction of continuously evolving graphs.

\section*{Acknowledgments}
This work was supported by National Natural Science Foundation of China (Grant No.62072077 and No.61602097).

\section{Broader Impact}
The methods described in this paper can potentially be harnessed to improve accuracy in any application of graph neural networks where it is more expensive or difficult to retrain the models frequently. For example, the graph-based recommender systems should make recommendations to a large number of new users. However, it is often prohibitive to continually update the online model to provide the most up-to-date recommendations. Our method offers a cost-efficient solution for such incremental recommender systems. In this spirit, this work may be beneficial to a range of applications requiring stable model performance but subjecting to limited computational resources. Although we anticipate a positive impact for applications where model predictions' robustness and stability are essential, e.g., in E-commerce and intelligent traffic systems, we also recognize possible malicious applications such as unauthorized surveillance and privacy intrusion.

\bibliography{AAAI20}


\end{document}

%% file: 1_Introduction.tex
Applying deep learning methods for graph data analytics tasks has recently generated a significant research interest~\cite{wu2019comprehensive}. Plenty of models have been developed to tackle various graph-related learning tasks, including node classification, link prediction, broader graph classifications, etc. Earlier efforts~\cite{perozzi2014deepwalk, tang2015line, grover2016node2vec} mainly focused on encoding nodes in networks/graphs\footnote{Whenever there is no ambiguity, the terms networks and graphs will be used interchangeably throughout this paper.} into a low-dimensional vector space, while preserving both the topological structure and the node attribute information in an unsupervised manner. However, researchers have recently shifted from developing sophisticated deep learning models on Euclidean-like domains (e.g., image, text) to non-Euclidean graph structure data. This, in turn, resulted in many notable Graph Neural Networks (GNNs) -- e.g., GCN~\cite{kipf2016semi}, GraphSAGE~\cite{hamilton2017inductive}, GAT~\cite{velivckovic2017graph}, SGC~\cite{wu2019simplifying}, and GIN~\cite{xu2018powerful}.

\begin{figure}[t]
	\centering
	\includegraphics[width=0.47\textwidth]{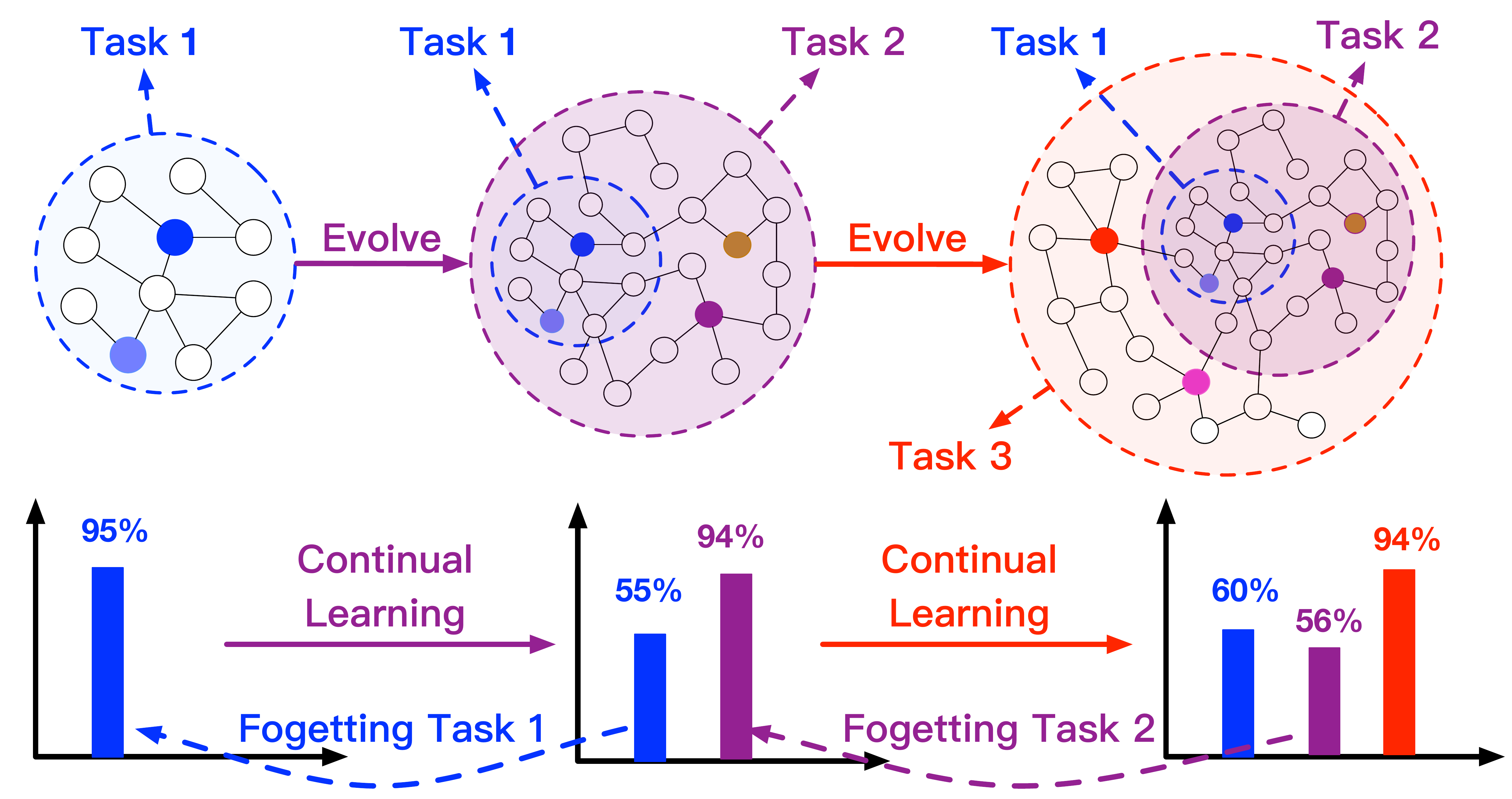}
	\caption{A CGL example. Nodes are papers and links indicate there is a citation between two papers. The task is to classify each node into several predefined classes (e.g., topics). From left to right is the evolution of the citation network. The histograms in different colors denote the node classification performance for each corresponding task.}
	\label{Problem}
\end{figure}

Despite significant breakthroughs achieved in GNNs, existing models -- in both static and dynamic graph settings -- primarily focus on a single task. Learning multiple tasks in sequence remains a fundamental challenge for GNNs. A natural question is how do these popular GNNs perform on learning a series of graph-related tasks, which is termed as \textit{Continual Graph Learning} (CGL) in this paper. Take Cora citation data as an illustrative scenario, as shown in Figure~\ref{Problem}. In the beginning, there are several nodes in the citation network representing some papers belong to a set of classes. We can train a node classification model on the current graph. However, the real-word citation network is naturally evolving overtime. Consequently, a group of new nodes from new classes (some are labeled and the others are unlabeled) will be added into the graph. We expect that the same model can classify the new nodes. Over time, multiple groups of nodes have been added to the network at different time-points. We term classifying each group of nodes as a task and expect to learn a common node classifier across all tasks, rather than one for the entire graph. Towards that, we train the classifier using each set of nodes in a sequential way. However, this kind of training process can easily lead to the phenomenon known as catastrophic forgetting (cf. the bottom of Figure~\ref{Problem}), where the classifier is updated and overwritten after learning a new task -- which is likely to result in a significant drop on the classification performance of previous tasks. Note that the classes/labels in one task are different from those in other tasks. Therefore, this learning process is often perceived as task-incremental learning~\cite{de2019continual}, which has practical value for many real-world graph-related applications, such as web-scale recommender systems~\cite{ying2018graph}, traffic condition predictions~\cite{Chen2019}, and protein design~\cite{ingraham2019generative}.

Continual learning, also referred to as lifelong learning, sequential learning, or incremental learning, has recently drawn significant research attention. Its objective is to gradually extend the acquired knowledge for future learning, which is very similar to human intelligence~\cite{chen2016lifelong}. Continual learning focuses on learning multiple tasks sequentially, targeting at two general goals: (i) learning a new task does not lead to catastrophic forgetting of former tasks~\cite{mccloskey1989catastrophic,goodfellow2013empirical} and (ii) the model can leverage knowledge from prior tasks to facilitate the learning of new tasks. Catastrophic forgetting is a direct outcome of a more general problem in neural networks, the so-called ``stability-plasticity'' dilemma~\cite{grossberg2012studies}. While stability indicates the preservation of previously acquired knowledge, plasticity refers to the ability to integrate new knowledge. This stability-plasticity trade-off is an essential aspect of both artificial and biological neural intelligent systems. Existing studies in continual learning mainly focus on image classification and reinforcement learning tasks, which have yielded several successful methods -- e.g., iCaRL~\cite{rebuffi2017icarl}, GEM~\cite{lopez2017gradient}, EWC~\cite{kirkpatrick2017overcoming}, SI~\cite{zenke2017continual}, LwF~\cite{li2017learning}, and PackNet~\cite{mallya2018packnet}. However, despite the extensive studies and promising results, there are surprisingly few works on CGL. The three major reasons are: (i) graph (non-Euclidean data) is not independent and identically distributed data; (ii) graphs can be irregular, noisy and exhibit more complex relations among nodes; and (iii) apart from the node feature information, the topological structure in graph plays a crucial role in addressing graph-related tasks.

To bridge this gap, in this work, we target at solving the continual learning problem for graph-structured data through formulating a continual node classification problem. We also conduct an empirical investigation of catastrophic forgetting in GNNs. To our knowledge, we are among the first to analyze graph data in such a sequential learning setting. We present a novel and general \textbf{\underline{E}}xperience \textbf{\underline{R}}eplay \textbf{GNN} framework (\textbf{ER-GNN}) which stores a set of nodes as experiences in a buffer and replays them in subsequent tasks, providing the capability of learning multiple consecutive tasks and alleviating catastrophic forgetting. For the experience selection, besides two intuitive strategies, we propose a novel scheme built upon influence function~\cite{hampel2011robust, koh2017understanding}, which performed quite favorably in our evaluation. In summary, we make the following contributions:
\begin{itemize}
	\item We present the continual graph learning (CGL) paradigm and formulate a new continual learning problem for node classification. The main difference from previous GNN works is that we aim to learn multiple consecutive tasks rather than a single task.
	
	\item We conduct an empirical investigation of the continual node classification task, demonstrating that existing GNNs are in the dilemma of catastrophic forgetting when learning a stream of tasks in succession.
	
	\item To address the catastrophic forgetting issue, we develop a generic experience replay based framework that can be easily combined with any popular GNNs model. Apart from two intuitive experience selection schemes, we propose a novel strategy based on influence function.
	
	\item We conduct extensive experimental evaluations using three benchmarks to demonstrate our framework's superiority over several state-of-the-art GNNs.
\end{itemize}

%% file: 2_RelatedWork.tex
\subsection{Graph Neural Networks}
Graph neural networks have recently become powerful models for learning representations of graphs. Many excellent GNN models have been proposed to exploit the structural information underlying graphs. They also potentially benefit many real-world applications, ranging from node classification and link prediction to traffic prediction and better recommendation~\cite{kipf2016semi,gao2018large,wu2019simplifying,zhang2019heterogeneous}. Most of the current GNNs can be categorized into two groups: spatial and spectral methods. Spatial methods aggregate the node representations directly from its neighborhood ~\cite{hamilton2017inductive,velivckovic2017graph, wang2019heterogeneous}, while the basic idea behind spectral approaches is to learn graph representation in the spectral domain where the learned filters are based on Fourier transformation~\cite{henaff2015deep,defferrard2016convolutional,kipf2016semi}. Although these GNNs have achieved great success in many graph-related applications, they only learn a single task. That is, they cannot be generalized to scenarios that require continuous learning while maintaining the model performance on previous tasks. In this work, we study a novel but fundamental problem in graph learning, i.e., how to train a GNN on a sequence of tasks, each of which is a typical graph-related problem such as node classification. Most importantly, the learned GNNs can successfully overcome the forgetting issue and allow us to retrospect earlier model behavior.

\subsection{Continual Learning}
Several approaches have been proposed to tackle catastrophic forgetting over the last few years. We can roughly distinguish three lines of work: (i) experience replay based methods; (ii) regularization-based methods; (iii) parameter isolation based methods. The first line of works stores samples in their raw format or compressed in a generative model. The stored samples from previous tasks are replayed when learning new tasks without significant forgetting. These samples/pseudo-samples can be used either for rehearsal -- approximating the joint training of previous and current tasks -- or to constrain the optimization~\cite{lopez2017gradient,rebuffi2017icarl,rolnick2019experience}. The second line of works proposes an additional regularization term in the loss function to consolidate previous knowledge when new data are available~\cite{kirkpatrick2017overcoming,li2017learning,zenke2017continual}. The last line of works attempts to prevent any possible forgetting of the previous tasks via models where different parameter subsets are dedicated to different tasks. When there is no constraint on the architecture's scale, it can be done by freezing the set of parameters learned after each previous task and growing new branches for new tasks. Alternatively, under a fixed architecture, methods proceed by identifying the parts used for the earlier tasks and masking them out during the training of the new task~\cite{mallya2018packnet,mallya2018piggyback,xu2018reinforced}. These methods have achieved great success in image classification and reinforcement learning tasks. However, they have not been investigated on graph-structured data, which motivates our study in this paper. Our method belongs to the family of experience replay based methods. Furthermore, we propose a new experience selection strategy based on influence function~\cite{hampel2011robust,koh2017understanding}, in addition to two intuitive experience selection schemes.

%% file: 3_Methodology.tex
This section describes the details of our proposed general framework ER-GNN for continual node classification. We begin with the formal definition of our problem and the biological theory of experience replay based methods, followed by the details of our ER-GNN where three experience selection strategies are presented.

\subsection{Problem Definition}
\label{Problem_Definition}
The settings of Continual Node Classification (i.e., task incremental learning) problem assume the existence of a collection of tasks: $ \mathcal{T} = \left\{ \mathcal{T}_{1}, \mathcal{T}_{2}, \dots, \mathcal{T}_{i}, \dots, \mathcal{T}_{M} \right\}$ which are encountered sequentially and each $\mathcal{T}_{i} \in \mathcal{T} (|\mathcal{T}| = M)$ is a node classification task.
Formally, the node classification task is defined as:
\begin{definition}[Node Classification]
	For each task $\mathcal{T}_{i}$, we have training node set $\mathcal{D}_{i}^\text{tr}$ and testing node set $\mathcal{D}_{i}^\text{te}$. Node classification aims to learn a task-specific classifier on $\mathcal{D}_{i}^\text{tr}$ that is excepted to classify each node in $\mathcal{D}_{i}^\text{te}$ into correct class $(y_{i}^{l} \in \mathcal{Y}_{i})$, where $\mathcal{Y}_{i} = \left\{y_{i}^{1}, y_{i}^{2}, \dots, y_{i}^{l}, \dots, y_{i}^{L} \right\}$ is the label set and $L$ is the number of classes in task $\mathcal{T}_{i}$.
\end{definition}

In the continual graph learning setting, instead of focusing on a single task $\mathcal{T}_{i}$, we need to learn a series of node classification task set $\mathcal{T}$. That is, our goal is to learn a model $f_{\bm{\theta}}$ parameterized by $\bm{\theta}$ that can learn these tasks successively. In particular, we expect the classifier $f_{\bm{\theta}}$ to not only perform well on the current task but also overcome catastrophic forgetting with respect to the previous tasks.

\subsection{Biological Theory}
\label{Biological_Theory}
Complementary Learning Systems (CLS) is a well-supported model of biological learning in human beings. It suggests that neocortical neurons learn with an algorithm that is prone to catastrophic forgetting. The neocortical learning algorithm is complemented by a virtual experience system that replays memories stored in the hippocampus to continually reinforce tasks that have not been recently performed~\cite{mcclelland1995there, kumaran2016learning}. The CLS theory defines the complementary contribution of the hippocampus and the neocortex in learning and memory, suggesting that there are specialized mechanisms in the human cognitive system for protecting consolidated knowledge. The hippocampal system exhibits short-term adaptation and allows for the rapid learning of new information, which will, in turn, be transferred and integrated into the neocortical system for its long-term storage.

\subsection{Experience Node Replay}
\label{Framework}
Inspired by the CLS theory, we propose a novel and general framework dubbed ER-GNN that selects and preserves experience nodes from the current task and replays them in future tasks. The framework of our ER-GNN is outlined in Algorithm~\ref{alg:algorithm1}.

\begin{algorithm}[t]
	\caption{Framework of our ER-GNN.}
	\label{alg:algorithm1}
	\textbf{Input}: Continual tasks $\mathcal{T}$: $\left\{ \mathcal{T}_{1}, \mathcal{T}_{2}, \dots, \mathcal{T}_{i}, \dots, \mathcal{T}_{M} \right\}$; Experience buffer: $\mathbb{B}$; Number of examples in each class added to $\mathbb{B}$: $e$.
	
	\textbf{Output}: Model $f_{\bm{\theta}}$ which can mitigate catastrophic forgetting of preceding tasks.
	\begin{algorithmic}[1]
	\STATE Initialize $\bm{\theta}$ at random;
	\WHILE{continual task $\mathcal{T}$ remains}
		\STATE Obtain training set $\mathcal{D}_{i}^\text{tr}$ from current task $\mathcal{T}_{i}$

		\STATE Extract experience nodes $B$ from experience buffer $\mathbb{B}$
		
		\STATE Compute loss function: $\mathcal{L}_{\mathcal{T}_{i}}^{\prime}(f_{\bm{\theta}}, \mathcal{D}_{i}^\text{tr}, B)$
		
		\STATE Compute optimal parameters:
		
		$\bm{\theta} = \arg \min _{\bm{\theta} \in \Theta} (\mathcal{L}_{\mathcal{T}_{i}}^{\prime}(f_{\bm{\theta}}, \mathcal{D}_{i}^\text{tr}, B))$
		
		\STATE Select experience nodes $ \mathcal{E} = \operatorname{Select}(\mathcal{D}_{i}^\text{tr}, e)$
			
		\STATE Add $\mathcal{E}$ to experience buffer: $\mathbb{B}$ = $\mathbb{B} \cup \mathcal{E}$	
		
		\STATE $\mathcal{T} = \mathcal{T} \setminus \{\mathcal{T}_i \} $
	\ENDWHILE
	\STATE \textbf{Return} model $f_{\bm{\theta}}$
	\end{algorithmic}
\end{algorithm}

When learning a task $\mathcal{T}_{i}$, we acquire its training set $\mathcal{D}_{i}^\text{tr}$ and testing set $\mathcal{D}_{i}^\text{te}$. Subsequently, we select examples $B$ from the experience buffer $\mathbb{B}$. Then we feed the training set $\mathcal{D}_{i}^\text{tr}$ and  the experience nodes $B$ together to our classifier $f_{\bm{\theta}}$. A natural loss function choice for node classification task is the cross-entropy loss function:

\begin{align}
\label{Loss}
\mathcal{L}_{\mathcal{T}_{i}}(f_{\bm{\theta}}, \mathcal{D}) = - & (\sum_{(\bm{x_{i}}, y_{i}) \in \mathcal{D}} (y_{i} \log f_{\bm{\theta}}(\bm{x_{i}}) 
\notag
\\& + (1-y_{i}) \log (1-f_{\bm{\theta}}(\bm{x_{i}})))).
\end{align}
Note that the number of nodes in training set $\mathcal{D}_{i}^\text{tr}$ is usually significantly larger than the size of the experience buffer. Here we need a weight factor $\beta$ to balance the influence from $\mathcal{D}_{i}^\text{tr}$ and $B$, averting the model from favoring a particular set of nodes. According to our empirical observations, we design a dynamic weight factor mechanism:
\begin{equation}
\label{Beta}
\beta = |B| / ({|\mathcal{D}_{i}^\text{tr}| + |B|}),
\end{equation}
where $|\mathcal{D}_{i}^\text{tr}|$ and $|B|$ are the number of nodes in the training set of $\mathcal{T}_{i}$ and the size of current experience buffer, respectively. The basic idea of this choice is to dynamically balance the following two losses:

\begin{align}
\label{Loss_1}
\mathcal{L}_{\mathcal{T}_{i}}^{\prime}(f_{\bm{\theta}}, \mathcal{D}_{i}^\text{tr}, B) = & \beta \mathcal{L}_{\mathcal{T}_{i}}(f_{\bm{\theta}}, \mathcal{D}_{i}^\text{tr}) 
\notag
\\&+ (1 - \beta) \mathcal{L}_{\mathcal{T}_{i}}(f_{\bm{\theta}}, B).
\end{align}

Subsequently, we perform parameter updates to obtain the optimal parameters by minimizing the empirical risk:
\begin{equation}
\bm{\theta} = \arg \min _{\bm{\theta} \in \Theta} (\mathcal{L}_{\mathcal{T}_{i}}^{\prime}(f_{\bm{\theta}}, \mathcal{D}_{i}^\text{tr}, B)),
\end{equation}
which can be trained with any optimization methods such as Adma optimizer~\cite{kingma2014adam}. After updating the parameters we need to select certain nodes in $\mathcal{D}_{i}^\text{tr}$ as experience nodes $\mathcal{E}$ and add them into the experience buffer $\mathbb{B}$. $ \mathcal{E} = \operatorname{Select}(\mathcal{D}_{i}^\text{tr}, e)$ means choosing $e$ nodes in each class as experiences of this task which will be cached into experience buffer $\mathbb{B}$. 

The experience selection strategy is crucial to the performance of CGL. We now turn our attention to the problem of identifying which nodes should be stored in the experience buffer $\mathbb{B}$. In the sequel, we present three schemes based on \textit{mean of feature}, \textit{coverage maximization}, and \textit{influence maximization}.

\noindent\textbf{Mean of Feature (MF)}: Intuitively, the most representative nodes in each class are the ones closest to the average feature vector. Similar to the prior work~\cite{rebuffi2017icarl} on continual image classification, for each task we compute a prototype for each class and choose $e$ nodes that are the first $e$ nodes closest to this prototype to form the experiences. In some attributed networks, each node has its own attribute vector $\bm{x_{i}}$ and embedding vector $\bm{h_{i}}$. This means we can obtain our prototypes based on the average attribute vector or the average embedding vector. Therefore, in the MF scheme, we compute the mean of attribute/embedding vector to produce a prototype and choose $e$ nodes whose attribute/embedding vectors are closest to the prototype:
\begin{equation}
\mathbf{c}_{l} = \frac{1}{|S_{l}|} \sum_{(\bm{x_{i}}, y_{i}) \in S_{l}} \bm{x_{i}},  \mathbf{c}_{l} = \frac{1}{|S_{l}|} \sum_{(\bm{x_{i}}, y_{i}) \in S_{l}} \bm{h_{i}},
\end{equation}
where $S_{l}$ is the set of training nodes in class $l$ and $\mathbf{c}_{l}$ is the prototype of nodes in class $l$. It is worth noting that although we can calculate prototypes from embedding vectors, we save the original nodes (i.e., $\bm{x_i}$) as our experiences since we will feed these nodes to our model again when learning new tasks.

\noindent\textbf{Coverage Maximization (CM)}: When the number of experience nodes $e$ in each class is small, it might be helpful to maximize the coverage of the attribute/embedding space. Drawing inspiration from the prior work~\cite{de2016improved} on continual reinforcement learning, we hypothesize that approximating a uniform distribution over all nodes from the training set $\mathcal{D}_{i}^\text{tr}$ in each task $\mathcal{T}_{i}$ can facilitate choosing experience nodes. To maximize the coverage of the attribute/embedding space, we rank the nodes in each class according to the number of nodes from other classes in the same task within a fixed distance $d$:
\begin{equation}
\mathcal{N}(v_{i}) = \left\{v_{j} | \text {dist}(v_{i}-v_{j})<d, \mathcal{Y}(v_{i}) \neq \mathcal{Y}(v_{j}) \right\},
\end{equation}
where $\mathcal{Y}(v_{i})$ is the label of node $v_{i}$, $\mathcal{N}(v_{i})$ is the set of nodes coming from different classes within distance $d$ to $v_{i}$. We can choose $e$ nodes with the lowest $|\mathcal{N}(v_{i})|$ in each class as our experiences. Similarly to MF, we can maximize the coverage of either the attribute space or the embedding space, but only store the original nodes as experience.

\noindent\textbf{Influence Maximization (IM)}: When training on each task $\mathcal{T}_{i}$, we can remove one training node $v_{\star}$ from the training set $\mathcal{D}_{i}^\text{tr}$ and obtain a new training set $\mathcal{D}_{i\star}^\text{tr}$. Then we can calculate the optimal parameters $\bm{\theta_{\star}}$ as: 
\begin{equation}
\bm{\theta_{\star}} = \arg \min _{\bm{\theta} \in \Theta}(\mathcal{L}_{\mathcal{T}_{i}}^{\prime}(f_{\bm{\theta}}, \mathcal{D}_{i\star}^\text{tr}, B)),
\end{equation}
resulting in a change in model optimal parameters: $\bm{\theta_{\star}} - \bm{\theta}$. However, obtaining the influence of every removed training node $v_{\star}$ is prohibitively expensive since it requires retraining the model for each removed node. Fortunately, influence function~\cite{hampel2011robust} provides theoretical foundations for estimating the change of parameters without model retraining and has been successfully used in previous works~\cite{koh2017understanding} for explaining the behaviors of neural networks. The basic idea is to compute the change of optimal parameters if $v_{\star}$ was upweighted by some small $\epsilon$, which gives the new parameters:
\begin{equation}
\label{redefine}
\bm{\theta_{\epsilon,\star}} \stackrel{\text{def}}{=} \arg \min _{\bm{\theta} \in \Theta}(\mathcal{L}_{\mathcal{T}_{i}}^{\prime}(f_{\bm{\theta}}, \mathcal{D}_{i}^\text{tr}, B) + \epsilon \mathcal{L}_{\mathcal{T}_{i}}(f_{\bm{\theta}}, v_{\star})),
\end{equation}
where the influence of upweighting $v_{\star}$ on the parameters $\bm{\theta}$ is given by (the details of the derivation can be found in Appendix A.1):
\begin{equation}
\label{derivation}
\mathcal{I}_{\mathrm{up}, \bm{\theta}}(v_{\star}) \stackrel{\text{def}}{=} \left. \frac{\partial \bm{\theta_{\epsilon,\star}}}{\partial \epsilon} \right|_{\epsilon=0} =-\mathbf{H}_{\bm{\theta}}^{-1} \nabla_{\bm{\theta}} \mathcal{L}_{\mathcal{T}_{i}}(f_{\bm{\theta}}, v_{\star}),
\end{equation}
where $\mathbf{H}_{\bm{\theta}}$ is the Hessian matrix that can be computed as:
\begin{equation}
\mathbf{H}_{\bm{\theta}} \stackrel{\text { def }}{=} \frac{1}{({|\mathcal{D}_{i}^\text{tr}| + |B|})} \sum_{j=1}^{({|\mathcal{D}_{i}^\text{tr}| + |B|})} \nabla_{\bm{\theta}}^{2} \mathcal{L}_{\mathcal{T}_{i}}(f_{\bm{\theta}}, v_{j}),
\end{equation}
and Eq.\eqref{redefine} suggests that removing node $v_{\star}$ is the same as upweighting it by $\epsilon=- (1 / ({|\mathcal{D}_{i}^\text{tr}| + |B|}))$. Thus, we can linearly approximate the parameter change of removing $v_{\star}$ as $\bm{\theta_{\star}} - \bm{\theta} \approx -(1 / ({|\mathcal{D}_{i}^\text{tr}| + |B|})) \mathcal{I}_{\mathrm{up}, \bm{\theta}}(v_{\star})$, without retraining the model. However, the Frobenius norm of $\bm{\theta_{\star}} - \bm{\theta}$ is usually too small to find the exact $\bm{\theta_{\star}}$. Besides, calculating the inverse of matrix $\mathbf{H}_{\bm{\theta}}$ is computationally expensive. To avoid these issues, we can alternatively estimate the influence of upweighting a training node $v_{\star}$ on the loss for a testing node $v_\text{test}$:

\begin{align}
\mathcal{I}_{\mathrm{up}, \operatorname{loss}}&(v_{\star}, v_\text{test})  \stackrel{\text {def}}{=}  \left. \frac{\partial \mathcal{L}_{\mathcal{T}_{i}}(f_{\bm{\theta_{\epsilon,\star}}}, v_{\text{test}})}{\partial \epsilon} \right|_{\epsilon=0} \notag
\\& =\nabla_{\bm{\theta}} \mathcal{L}_{\mathcal{T}_{i}}(f_{\bm{\theta}}, v_{\text{test}}) \left.\frac{\partial \bm{\theta_{\epsilon,\star}}}{\partial \epsilon} \right|_{\epsilon=0} \notag
\\& =-\nabla_{\bm{\theta}} \mathcal{L}_{\mathcal{T}_{i}}(f_{\bm{\theta}}, v_{\text{test}})^{\mathrm{T}} \mathbf{H}_{\bm{\theta}}^{-1} \nabla_{\bm{\theta}} \mathcal{L}_{\mathcal{T}_{i}}(f_{\bm{\theta}}, v_{\star}).
\end{align}
Therefore, we can sum all testing nodes' influence $\mathcal{I}_{\mathrm{up}, \operatorname{loss}}(v_{\star}, v_\text{test})$ to derive the influence of the training node $v_{\star}$. During this process, one can use implicit Hessian-vector products (HVPs) to approximate $w_{\text {test}} \stackrel{\text{def}}{=} \mathbf{H}_{\bm{\theta}}^{-1} \nabla_{\bm{\theta}} \mathcal{L}_{\mathcal{T}_{i}}(f_{\bm{\theta}}, v_{\text{test}})^{\mathrm{T}}$~\cite{koh2017understanding}, i.e., $\mathcal{I}_{\mathrm{up}, \operatorname{loss}}(v_{\star}, v_\text{test}) = -w_{\text{test}}\nabla_{\bm{\theta}} \mathcal{L}_{\mathcal{T}_{i}}(f_{\bm{\theta}}, v_{\star})$, so as to speed up the computation. Since the Hessian $\mathbf{H}_{\bm{\theta}}$ is positive semi-definite by assumption, we have:

\begin{equation}
w_{\text{test}} \equiv \arg \min _{\alpha}\left\{\frac{1}{2} \alpha^{\mathrm{T}} \mathbf{H}_{\bm{\theta}} \alpha- \nabla_{\bm{\theta}} \mathcal{L}_{\mathcal{T}_{i}}(f_{\bm{\theta}}, v_{\text{test}})^{\mathrm{T}} \alpha\right\}, \notag
\end{equation}
where the exact solution $\alpha$ can be obtained with conjugate gradients that only requires the evaluation of $\mathbf{H}_{\bm{\theta}} \alpha$ instead of explicitly computing $\mathbf{H}_{\bm{\theta}}^{-1}$. 

We hypothesize that the larger the influence of $v_{\star}$, the more representative $v_{\star}$ for this task. Thus, we choose the first $e$ representative nodes in each class as our experiences. To our knowledge, we are the first to incorporate influence function into continual learning settings to guide the selection of experience samples. We also study the effectiveness of the influence function based experience selection scheme in our experiments. The empirical results verify the generalization of this strategy.

Our framework ER-GNN does not impose \textit{any} restriction on GNNs architecture and can be easily incorporated into most of the current GNN models. In our evaluation, we implement our ER-GNN with a vanilla GAT~\cite{velivckovic2017graph}, forming an instance of our framework -- ER-GAT.

%% file: 4_Experiments.tex
We now present the results from the empirical evaluation of our framework for continual node classification tasks to demonstrate its effectiveness and applicability. For reproducibility, the source code and datasets are provided in the supplementary materials. We begin with systematically investigating to what extent the state-of-the-art GNNs forget on learning a sequence of node classification tasks, followed by the performance lift of our ER-GNN. Subsequently, we verify the applicability of our ER-GNN and study the hyperparameter sensitivity of our model. We also conduct an extra experiment to show that our IM-based experience selection scheme can be generalized to tasks beyond CGL.

\noindent\textbf{Datasets}: To evaluate the performance of our model on solving the CGL problem, we conduct experiments on three benchmark datasets: Cora~\cite{sen2008collective}, Citeseer~\cite{sen2008collective}, and Reddit~\cite{hamilton2017inductive} that are widely used for evaluating the performance of GNN models. To meet the requirements of the continual graph learning (task-incremental) setting, we construct $3$ tasks on Cora and Citeseer, and each task is a $2$-way node classification task, i.e., there are $2$ classes in each task. For Reddit, we generate $8$ tasks and each task is a $5$-way node classification task due to its relatively large number of nodes and unique labels. We note that each task is a new task since classes in different tasks are entirely different. The statistics of the datasets and continual task settings are shown in Table~\ref{Datasets}.
\begin{table}[t] 
	\centering
	\begin{tabular}{r|ccc}
		\hline
		&Cora &Citeseer &Reddit \\
		\hline
		\text{\#} Nodes 	&2,708 &3,327 &232,965\\
		\text{\#} Node Attributes 	&1,433 &3,703 &602\\
		\text{\#} Total Classes 	&7 &6 &41\\
		\hline
		\text{\#} Tasks	&3 &3 &8\\
		\text{\#} Classes in Each Task &2 &2 &5\\
		\hline
	\end{tabular}
	\caption{Descriptive statistics and task settings of three datasets.}
	\label{Datasets}
\end{table}

\begin{table*}[t]
	\renewcommand{\multirowsetup}{\centering}
	\centering
	{\begin{tabular}{c|cc|cc|cc}
			\hline
			Datasets& \multicolumn{2}{c|}{Cora} & \multicolumn{2}{c|}{Citeseer} & \multicolumn{2}{c}{Reddit} \\
			\hline
			\diagbox[]{Methods}{Metrics}& PM & FM & PM & FM & PM & FM \\
			\hline
			DeepWalk  & 85.63\% & 34.51\% & 64.79\% & 25.92\% & 76.93\% & 33.24\% \\
			Node2Vec  & 85.99\% & 35.46\% & 65.18\% & 24.87\% & 78.24\% & 34.66\% \\
			GraphSAGE & 94.15\% & 37.73\% & 81.26\% & 28.06\% & 95.01\% & 40.06\% \\
			GIN       & 90.17\% & 33.81\% & 74.92\% & 27.42\% & 93.75\% & 36.28\% \\
			GCN       & 93.62\% & 31.90\% & 80.63\% & 25.47\% & 94.43\% & 35.17\% \\
			SGC       & 93.06\% & 33.93\% & 78.18\% & 28.31\% & 94.01\% & 38.59\% \\
			GAT       & 94.19\% & 30.84\% & 81.37\% & 25.06\% & 95.13\% & 34.97\% \\
			\hline
			ER-GAT-Random & 93.58\% & 29.17\% & 81.48\% & 23.73\% & 93.84\% & 32.79\% \\
			\hline
			ER-GAT-MF & 94.15\% & 22.49\% & 80.03\% & 17.96\% & 94.18\% & 26.44\% \\
			$\text{ER-GAT-MF}^{\star}$ & 94.23\% & 21.88\% & \textbf{81.83\%} & 17.83\% & 94.63\% & 23.54\% \\
			ER-GAT-CM & 93.98\% & 22.14\% & 78.78\% & 18.03\% & 93.33\% & 26.17\% \\
			$\text{ER-GAT-CM}^{\star}$ & 94.25\% & \textbf{21.03\%} & 80.86\% & 17.86\% & 94.23\% & 23.15\% \\
			ER-GAT-IM & \textbf{95.66\%} & 21.14\% & 80.85\% & \textbf{17.08\%} & \textbf{95.36\%} & \textbf{23.09\%} \\
			\hline
	\end{tabular}}
	\caption{Performance comparisons among algorithms. The bold values denote the best performance.}
	\label{Performance}
\end{table*}

\noindent\textbf{Baselines}: To demonstrate the effectiveness of our proposed framework, we compare ER-GNN with the following GNNs for continual node classification tasks:

\noindent $\bullet$ \textbf{Deepwalk}~\cite{perozzi2014deepwalk}: Deepwalk uses local information from truncated random walks as input to learn a representation which encodes structural regularities.

\noindent $\bullet$ \textbf{Node2Vec}~\cite{grover2016node2vec}: Node2Vec learns a mapping of nodes to a low-dimensional space of features that maximize the likelihood of preserving network neighborhoods of nodes.

\noindent $\bullet$ \textbf{GCN}~\cite{kipf2016semi}: GCN uses an efficient layer-wise propagation rule based on a first-order approximation of spectral convolution on the graph.

\noindent $\bullet$ \textbf{GraphSAGE}~\cite{hamilton2017inductive}: GraphSAGE learns a function that generates embeddings by sampling and aggregating features from the node’s local neighborhood.

\noindent $\bullet$ \textbf{GAT}~\cite{velivckovic2017graph}: GAT introduces an attention-based architecture to perform node classification of graph-structured data. The idea is to compute each node's hidden representations by attending over its neighbors, following a self-attention strategy.

\noindent $\bullet$ \textbf{SGC}~\cite{wu2019simplifying}: SGC reduces this complexity in GCN through successively removing nonlinearities and collapsing weight matrices between consecutive layers.

\noindent $\bullet$ \textbf{GIN}~\cite{xu2018powerful}: GIN develops a simple architecture that is probably the most expressive among the class of GNNs and is as powerful as the Weisfeiler-Lehman graph isomorphism test.

\noindent\textbf{Experimental Setting}: We implement ER-GNN with GAT, forming an example of our framework -- ER-GAT, together with our three experience selection strategies. In ER-GAT, we can readily obtain the embedding vector (before the last softmax layer). Thus, we get ER-GAT-MF, $\text{ER-GAT-MF}^{\star}$, ER-GAT-CM, $\text{ER-GAT-CM}^{\star}$, and ER-GAT-IM, where -MF, -$\text{MF}^{\star}$, -CM, -$\text{CM}^{\star}$, and -IM represent the mean of the attributes, the mean of embeddings, attribute space coverage maximization, embedding space coverage maximization, and influence maximization, respectively. The settings of all baselines and the network architecture (i.e., GAT) in our implementation are the same as suggested in the respective original papers. Without otherwise specified, we set the number of experiences stored in the experience buffer from each class as 1 (i.e., $e = 1$). However, we note that a larger value of $e$ would result in better performance.

\noindent\textbf{Metric}: To measure the performance in the continual graph learning setup, we use performance mean (PM) and forgetting mean (FM) as the evaluation metrics~\cite{chaudhry2018riemannian}. Taking the Cora dataset as an example, when learning $3$ tasks sequentially, there are $3$ accuracy values, i.e., one for each task after learning this task, and $3$ forgetting values, i.e., the difference between the performance after learning a particular task and the performance after learning its subsequent tasks. The evaluation on Citeseer is the same as Cora. However, we use Micro F1 score as the performance metric instead of accuracy due to the imbalance between the nodes in different classes in the Reddit dataset. That is, PM and FM in Reddit represent the average Micro F1 Score and the average difference in Micro F1 Score, respectively.

\noindent \textbf{Catastrophic Forgetting in GNNs.} 
We first systematically evaluate the extent of catastrophic forgetting in current GNNs. Table~\ref{Performance} shows the results of performance comparison on node classification, from which we can observe that all GNNs as well as two graph embedding models suffer from catastrophic forgetting problems on previous tasks. For example, the FM value on Cora, Citeseer, and Reddit are 30+\%, 24+\%, and 32+\%, respectively. Interestingly, in some cases (e.g., Reddit), DeepWalk and Node2Vec perform better than GNNs in terms of FM, although their classification performance (i.e., PM) is lower compared to the GNNs. DeepWalk and Node2vec use truncated random walks for sampling node sequences. Once the number of sampled node sequences is large enough, the obtained node embeddings are more robust for continuous task learning, although their performance is not comparable to others on single-task learning. Therefore, it seems DeepWalk (or Node2Vec) sacrifices the performance on learning new tasks (i.e., plasticity) to relieve the catastrophic forgetting (i.e., stability) issue.

We also find some impressive results by examining the performance of different GNNs. First, as a simplified GCN, SGC is significantly faster than other GNN models because it removes the nonlinearity between the layers in GNN -- it only keeps the final one. We speculate that nonlinearity in-between multi-layers in GNN can help the model remember knowledge from previous tasks. Besides, GraphSAGE achieves higher PM, but it is prone to previous task forgetting, mainly because it uses pooling operation (e.g., mean here) for feature aggregation, which, however, may smooth unique node features in previous tasks. GIN is mediocre on Cora and Citeseer but performs well on Reddit. The reason for that is because GIN may suffer from overfitting issues on relatively small datasets, as observed in~\cite{wu2019simplifying}. In contrast, GAT performs well in terms of both PM and FM, which suggests that the attention mechanism is beneficial for both learning a new task and to a certain extent resisting catastrophic forgetting in continual graph-related task learning.

\noindent \textbf{Performance of ER-GNN.} Next, we compare our framework ER-GNN against other graph learning baselines. In addition to the three proposed node selection strategies, we also implement a random node selection scheme, called ER-GAT-Random, which randomly selects the experience nodes. Note that we report the average results of 10 runs. Table~\ref{Performance} shows the results of $3$ different experience selection strategies. Our framework successfully decreases the FM values by a significant margin without losing the ability to learn new tasks. The IM strategy performs favorably, which proves our motivation to exploit influence function for experience replay. Another expected finding is that $\text{ER-GAT-MF}^{\star}$ and $\text{ER-GAT-CM}^{\star}$ consistently outperform ER-GAT-MF and ER-GAT-CM, which indicates the more discriminative representations in the embedding space than the attribute space.

An important observation is that our ER-GAT performs comparably with vanilla GAT in terms of PM, and in some cases, our ER-GAT even outperforms the original GAT. This result implies that our approach does not sacrifice the plasticity since we expand the training set by augmenting the experience buffer nodes from previous tasks. This property is appealing as it resembles our human intelligence, i.e., we humans can not only remember previous tasks but also exploit the knowledge from preceding tasks to facilitate learning future tasks. 

\begin{figure}[t]
	\centering
	\subfigure[Cora]{
		\includegraphics[width=0.22\textwidth]{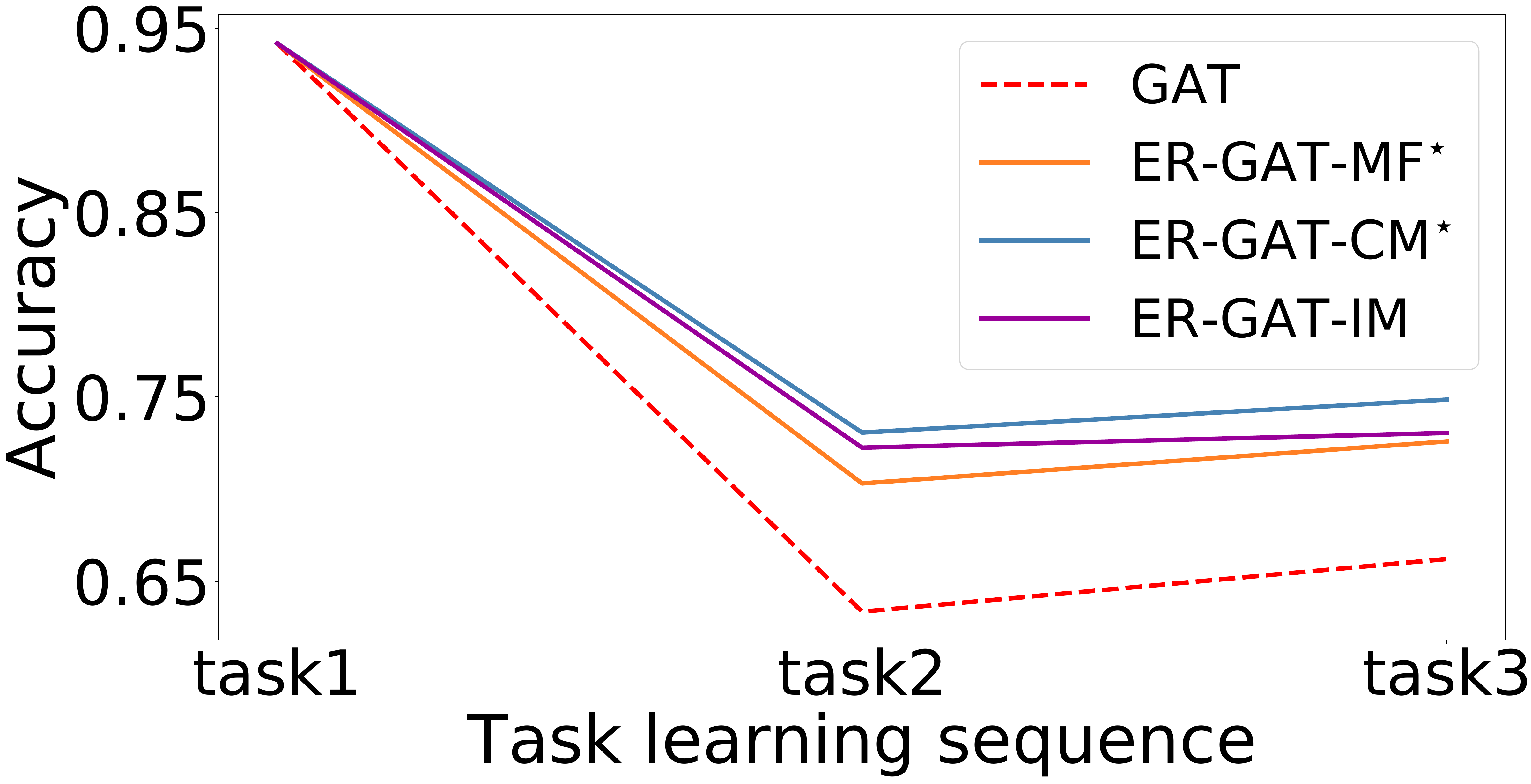}
		\label{Result1_Cora}
	}
	\subfigure[Citeseer]{
		\includegraphics[width=0.22\textwidth]{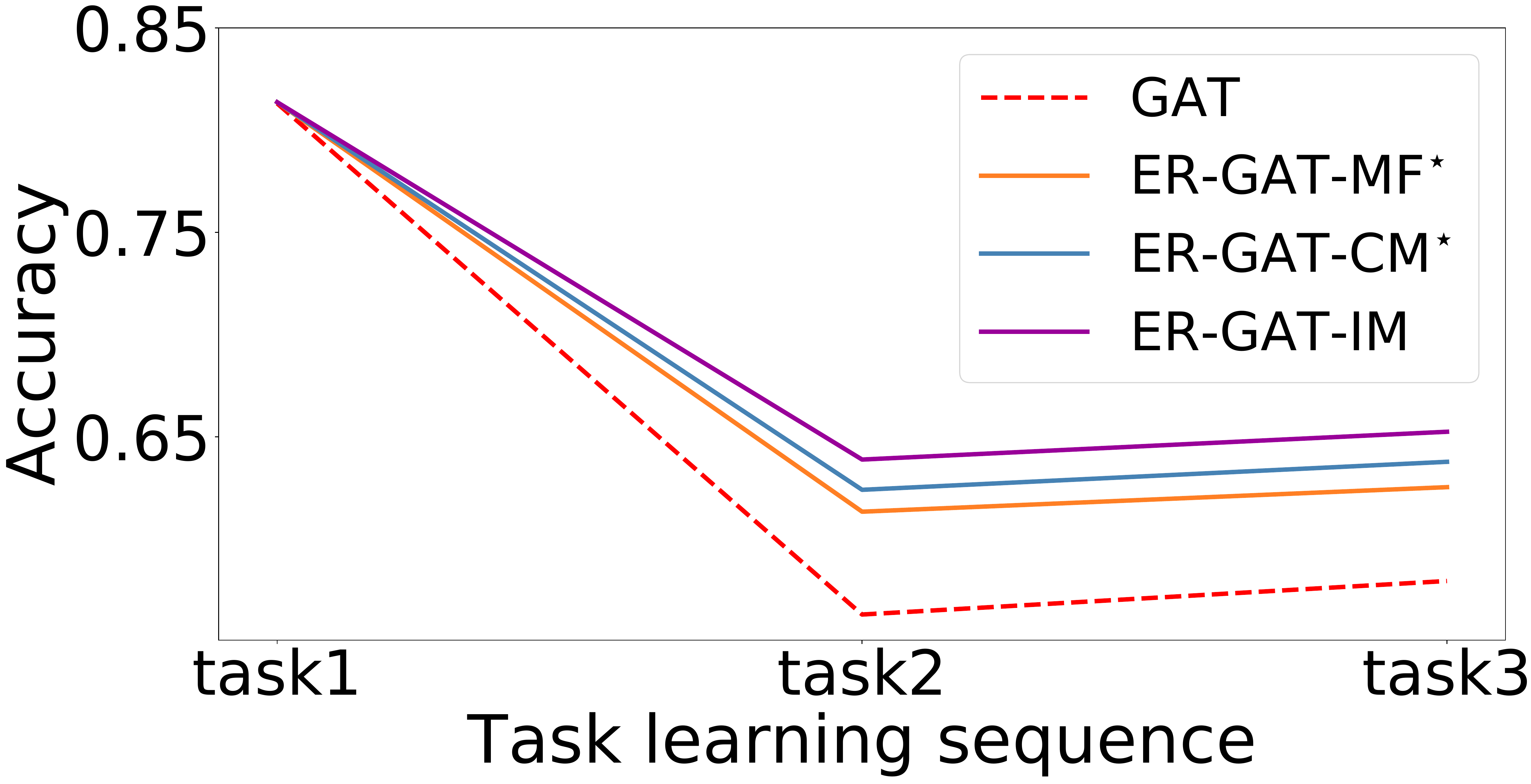}
		\label{Result1_Citeseer}
	}

	\subfigure[Reddit]{
		\includegraphics[width=0.23\textwidth]{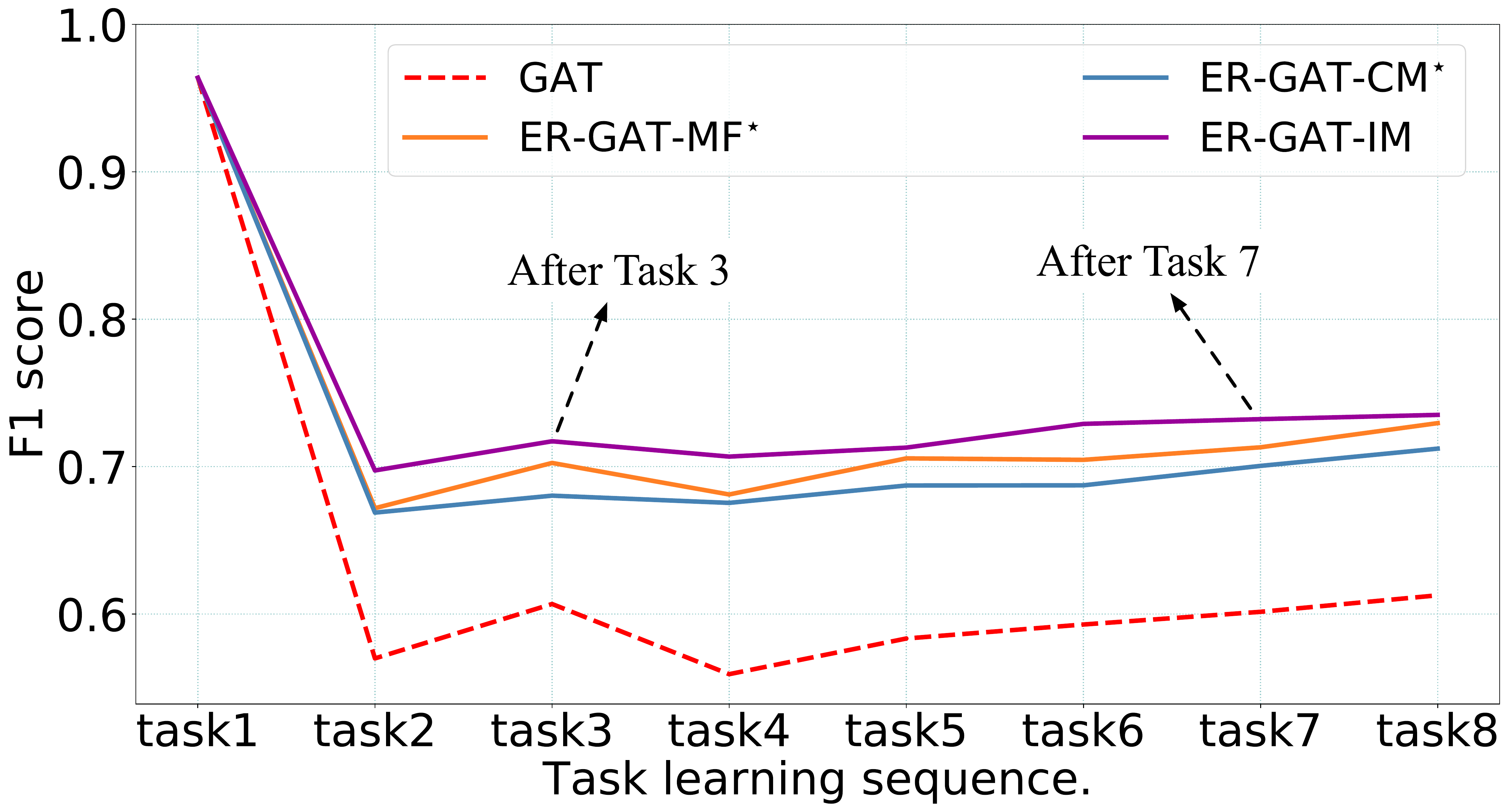}
		\label{Result1_Reddit}
	}
	\caption{Performance of PM evolved with the tasks.}
	\label{Result}
\end{figure}

To provide further insight into our approach, we plot the performance evolution of the models along with the increased tasks on three benchmarks in Figure~\ref{Result}. For clarity, we only plot GAT's results as it performs best among baselines, with the same reason for omitting ER-GAT-MF and ER-GAT-CM. From Figure~\ref{Result}, we can observe that our framework alleviates forgetting by a large margin compared with the original GAT. 

\noindent\textbf{Applicability of ER-GNN.} To demonstrate the applicability of our method in other GNNs, we instantiate our experience selection schemes with SGC and GIN, forming several variants of ER-SGC and ER-GIN. As shown in Table~\ref{Performance2}, ER-SGC and ER-GIN reduce the degree of forgetting in SGC and GIN by a large margin, which demonstrates the applicability of our approaches. Besides, we also observe that the IM-based models usually achieves the lowest FM results.

\begin{table}[t]
	\centering
	{\begin{tabular}{c|c|c|c}
		\hline
		\diagbox{Model}{Datasets} & Cora & Citeseer & Reddit\\
		\hline
		ER-SGC-MF & \textbf{25.00\%} & 20.34\% & 27.62\% \\
		ER-SGC-CM & 25.46\% & 19.38\% & 26.27\% \\
		ER-SGC-IM & 26.11\% & \textbf{17.16\%} & \textbf{25.04\%} \\
		\hline
		ER-GIN-MF & 27.98\% & 21.01\% & 25.36\% \\
		ER-GIN-CM & 27.38\% & 20.72\% & 24.23\% \\
		ER-GIN-IM & \textbf{26.74\%} & \textbf{20.68\%} & \textbf{23.75\%} \\
		\hline	
	\end{tabular}}
	\caption{Forgetting mean of ER-SGC and ER-GIN.}
	\label{Performance2}
\end{table}

\begin{figure}[t]
	\centering
	\subfigure[Cora]{
		\includegraphics[width=0.14\textwidth]{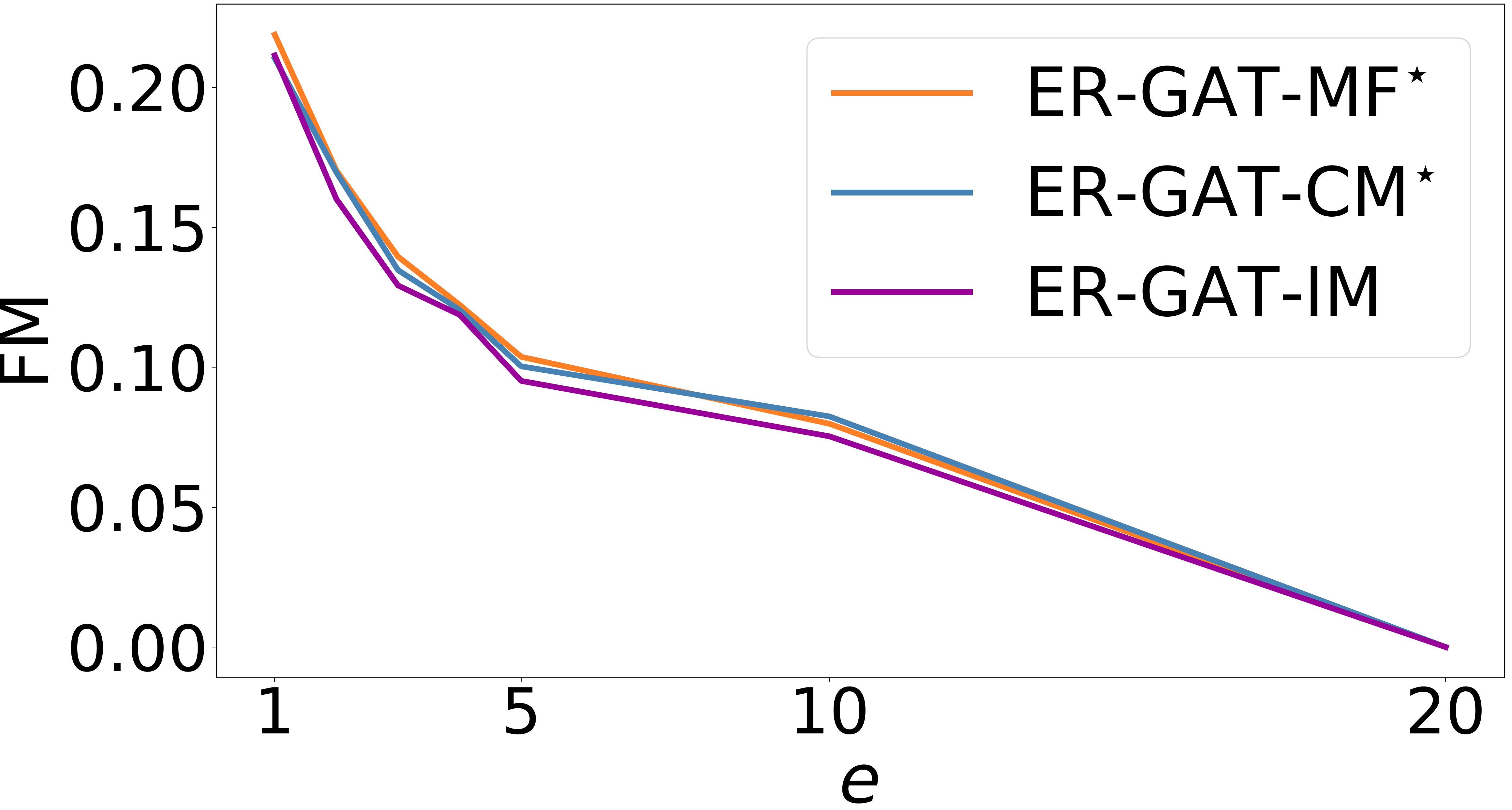}
		\label{Hyper_Parameter_Cora}
	}
	\subfigure[Citeseer]{
		\includegraphics[width=0.14\textwidth]{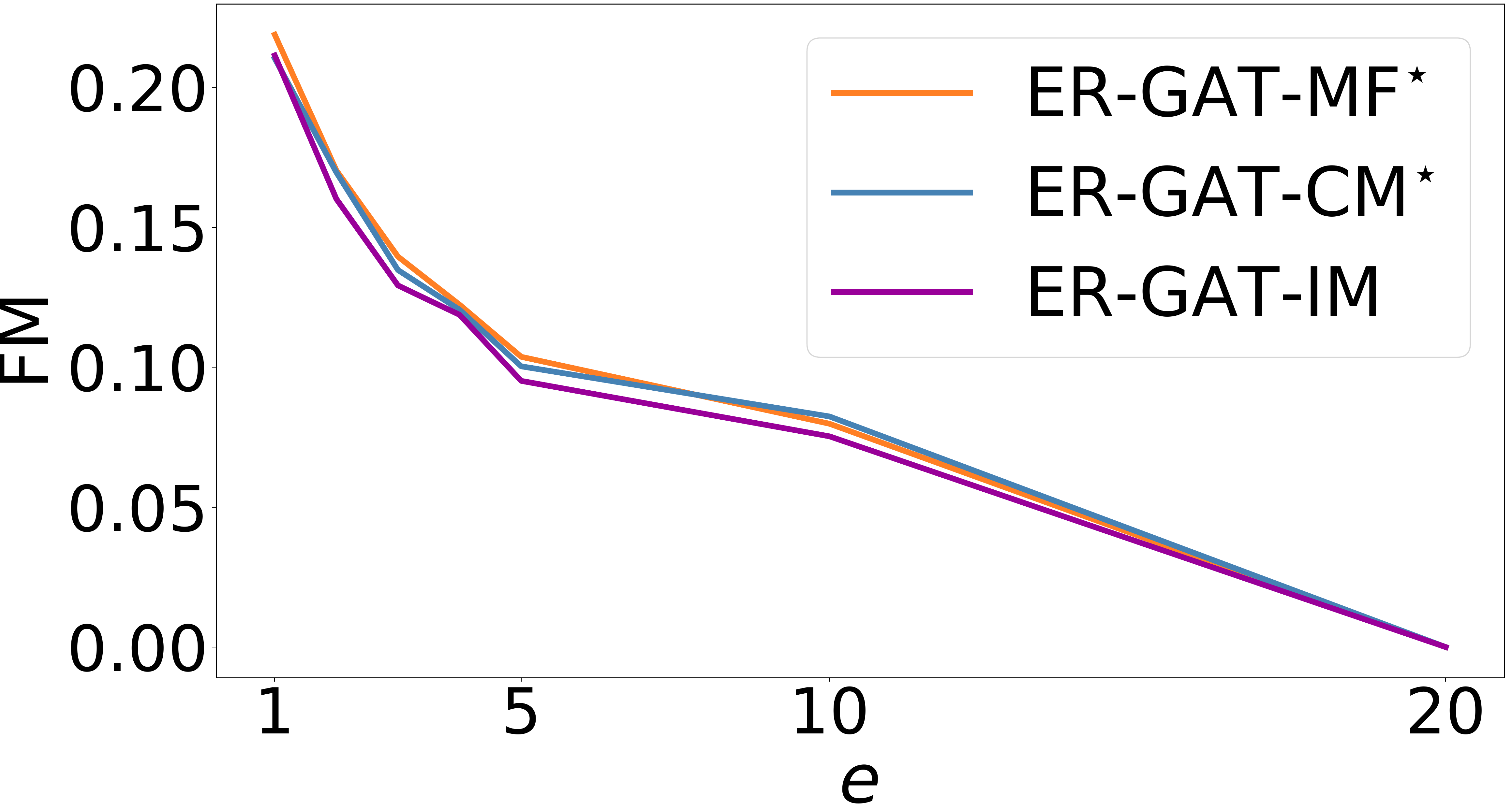}
		\label{Hyper_Parameter_Citeseer}
	}
	\subfigure[Reddit]{
		\includegraphics[width=0.14\textwidth]{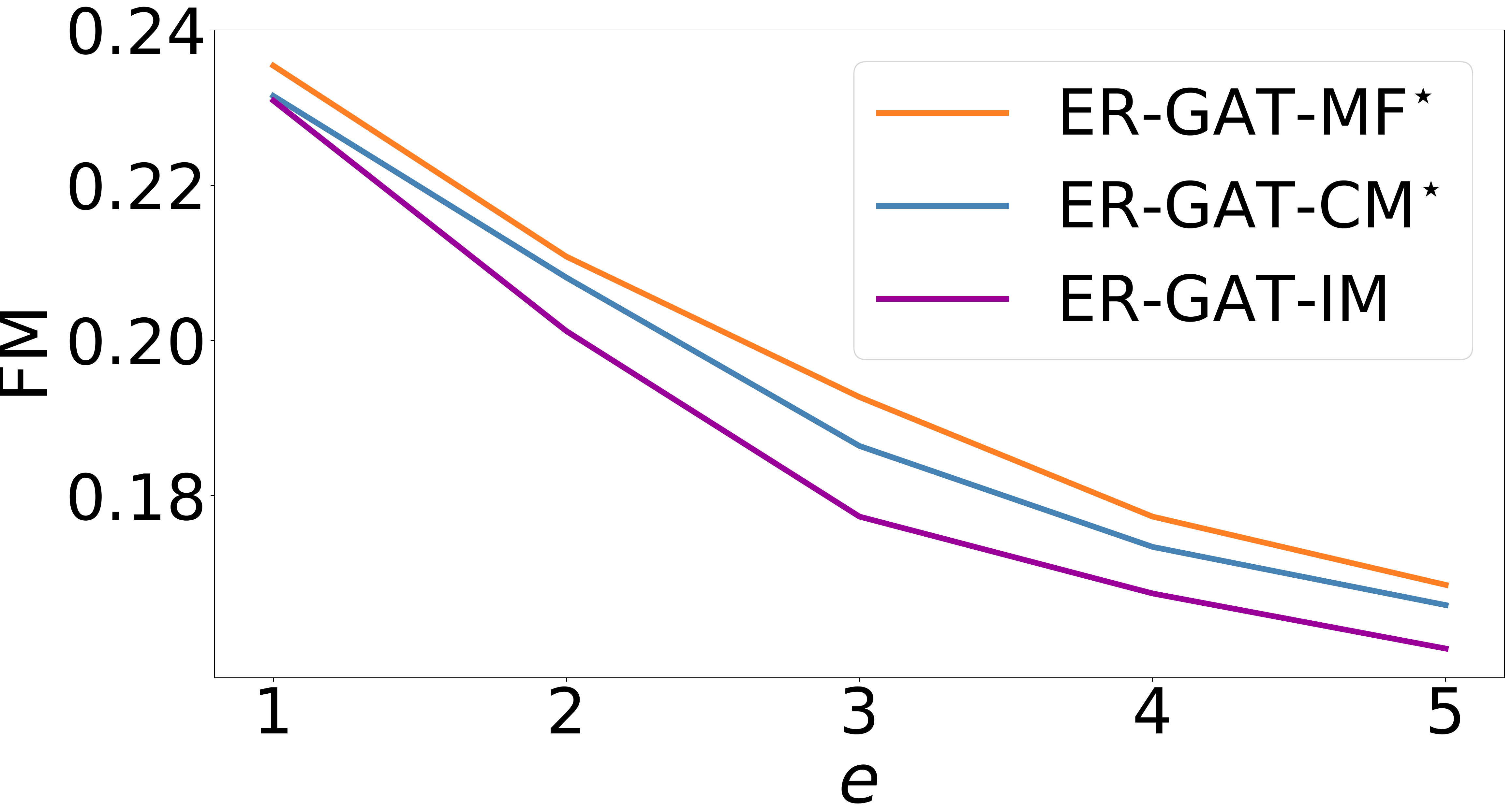}
		\label{Hyper_Parameter_Reddit}
	}
	\caption{The influence of $e$.}
	\label{Hyper_Parameter}
\end{figure}

\noindent\textbf{Influence of $e$.} 
The number of nodes stored in the buffer from each class is among the most crucial hyperparameters. We investigate its effect and present the results in Figure~\ref{Hyper_Parameter}. As expected, the more nodes stored in the buffer, the better the performance on alleviating the catastrophic forgetting. Furthermore, we can see that our model would not forget if we keep all training nodes in previous tasks -- note that there are 20 training nodes in each class on Cora and Citeseer dataset.

\noindent\textbf{Effect of IM on Continual Image Classification.} To verify the generalization of our influence function based experience selection strategy beyond CGL, we conduct an extra evaluation on a commonly used benchmark in continual image learning, i.e., Permuted MNIST dataset~\cite{kirkpatrick2017overcoming}. It is a variant of MNIST dataset of handwritten digits where each task has a certain random permutation of input pixels that applied to all the images of that task. The Permuted MNIST benchmark consists of $23$ tasks, and we use the forgetting mean (FM) as our evaluation metric. We use a fully-connected network with two hidden layers of 256 ReLU units each for all methods. The results are shown in Table~\ref{MNIST}, where FINETUNE means that the model is trained continually without any regularization and task memory suggested by~\cite{lopez2017gradient}. A new task's parameters are initialized from the previous task's parameters and can be considered a lower bound. The comparison between our approach and two well-known baselines (i.e., EWC~\cite{kirkpatrick2017overcoming} and GEM~\cite{lopez2017gradient}) are shown in Table~\ref{MNIST}, where we can see that our scheme is superior on overcoming the catastrophic forgetting. Our method and GEM -- experience reply based methods -- outperform EWC that is a regularization-based, indicating that extra experience buffer is more effective in overcoming catastrophic forgetting. Furthermore, our approach outperforms GEM, indicating that the nodes selected by IM are more representative than those chosen by gradients, i.e., the IM is more indicative than gradients when estimating the importance of a node.

\begin{table}[t] 
	\centering
	\begin{tabular}{c|c}
		\hline
		Method &Forgetting Mean \\
		\hline
		FINETUNE &0.29 \\
		EWC~\cite{kirkpatrick2017overcoming} &0.22 \\
		GEM~\cite{lopez2017gradient} &0.18\\
		\hline
		Ours &\textbf{0.14}\\
		\hline
	\end{tabular}
	\caption{The performance comparisons between different continual learning methods on Permuted MNIST data.}
	\label{MNIST}
\end{table}